%% file: main.tex
\pdfoutput=1
\def\authorversion{}

\documentclass[10pt,twocolumn,letterpaper]{article}

\ifdefined\authorversion
  \usepackage[pagenumbers]{cvpr}
\else
  \usepackage[pagenumbers]{cvpr}
\fi

\input{preamble}

\definecolor{cvprblue}{rgb}{0.21,0.49,0.74}
\usepackage[pagebackref,breaklinks,colorlinks,allcolors=cvprblue]{hyperref}

\title{Trust the Ray, Not the Depth:\\Ray-Lateral Agreement for Multi-View 3D Human Pose Fusion}

\author{Farnoosh Koleini$^{1,2}$\qquad Ahmed Helmy$^{1}$\qquad Pu Wang$^{1}$\qquad Xiwu Cao$^{2}$\\[0.5em]
{\normalsize $^{1}$University of North Carolina at Charlotte\qquad $^{2}$Canon U.S.A., Inc.}\\[0.2em]
{\tt\small fkoleini@dis.cusa.canon.com\qquad xcao@cusa.canon.com}\\
{\tt\small \{fkoleini, ahmed.helmy, pu.wang\}@charlotte.edu}}
\ifdefined\authorversion
  \hypersetup{pdftitle={Trust the Ray, Not the Depth: Ray-Lateral Agreement for Multi-View 3D Human Pose Fusion},
              pdfauthor={Farnoosh Koleini, Ahmed Helmy, Pu Wang, Xiwu Cao}}
\fi
\vspace{-2 mm}
\input{supp_labels}
\begin{document}
\maketitle
\input{sec/0_abstract}
\begin{figure}[!t]
  \centering
  \includegraphics[width=0.85\linewidth]{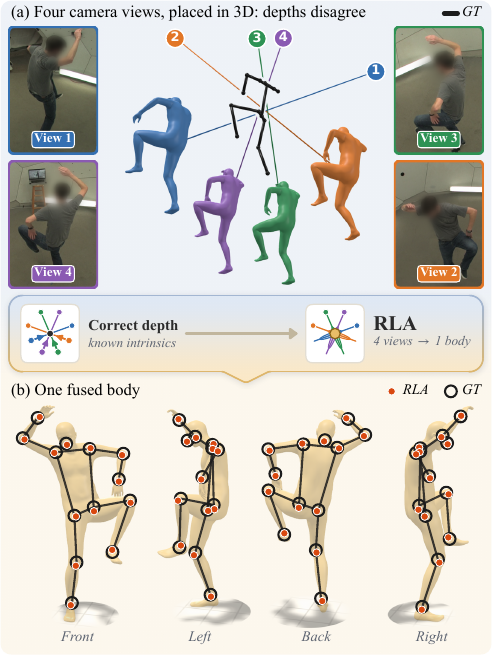}
  \caption{\textbf{Monocular 3D reconstructions from calibrated cameras disagree in depth; \method aligns and fuses them into a single body.}
  (CMU Panoptic; faces blurred.)
  (a)~Monocular estimates from four calibrated camera views, placed in a common coordinate system using the known camera calibration. Each reconstruction is displaced from the ground truth (GT), primarily along its camera ray, due to a mismatch between the network's assumed focal length and the camera's actual focal length.
  (b)~After correcting the depth mismatch, \method fuses the estimates point by point. Viewed from four directions, the fused joints closely align with GT.}
  \label{fig:teaser}
\end{figure}
\vspace{-2mm}
\vspace{-12pt}
\input{sec/1_intro}
\input{sec/2_related}
\input{sec/3_method}
\input{sec/4_experiments}
\input{sec/5_conclusion}
{
    \small
    \bibliographystyle{ieeenat_fullname}
    \bibliography{main}
}

\end{document}

%% file: preamble.tex
\newcommand{\TODO}[1]{\textbf{\color{red}[TODO: #1]}}

\newcommand{\missingnum}[1]{\textbf{\color{red}??}}
\ifdefined\submission
  \renewcommand{\TODO}[1]{\PackageError{paper}{TODO left in the text: #1}{}}

  \renewcommand{\missingnum}[1]{\PackageError{paper}{number not generated: #1}{}}
\fi
\ifdefined\authorversion
  \renewcommand{\TODO}[1]{}

  \AtBeginDocument{\setlength{\bibsep}{0pt}%
    \setlength{\textfloatsep}{14pt plus 2pt minus 2pt}\setlength{\dbltextfloatsep}{14pt plus 2pt minus 2pt}}
\fi

\usepackage{microtype}
\usepackage{colortbl}    %

\DeclareMathOperator*{\argmin}{arg\,min}
\DeclareMathOperator*{\median}{median}

\makeatletter
\renewcommand\paragraph{\@startsection{paragraph}{4}{\z@}{1.2ex \@plus .6ex \@minus .2ex}{-1em}{\normalfont\normalsize\bfseries}}
\makeatother

\newcommand{\method}{RLA\xspace}

\newcommand{\pnum}[4]{\ifcsname pn@#1@#2@#3@#4\endcsname\csname pn@#1@#2@#3@#4\endcsname\else\missingnum{pn@#1@#2@#3@#4}\fi}
\newcommand{\pinfo}[2]{\ifcsname pn@#1@#2\endcsname\csname pn@#1@#2\endcsname\else\missingnum{pn@#1@#2}\fi}
\InputIfFileExists{tables/panoptic_numbers}{}{}
\newcommand{\gnum}[2]{\ifcsname #1@#2\endcsname\csname #1@#2\endcsname\else\missingnum{#1@#2}\fi}
\InputIfFileExists{tables/panoptic_depth}{}{}
\InputIfFileExists{tables/panoptic_2d}{}{}
\InputIfFileExists{tables/panoptic_pairs}{}{}
\InputIfFileExists{tables/panoptic_ransac}{}{}
\InputIfFileExists{tables/panoptic_failures}{}{}
\InputIfFileExists{tables/panoptic_gold}{}{}
\InputIfFileExists{tables/panoptic_ablation_numbers}{}{}
\InputIfFileExists{tables/panoptic_runtime}{}{}
\InputIfFileExists{tables/panoptic_inference}{}{}
\InputIfFileExists{tables/panoptic_correction_numbers}{}{}
\InputIfFileExists{tables/panoptic_bones_numbers}{}{}
\InputIfFileExists{tables/panoptic_supp_numbers}{}{}
\InputIfFileExists{tables/panoptic_estimators}{}{}
\InputIfFileExists{tables/qualitative_numbers}{}{}
\InputIfFileExists{tables/panoptic_detector}{}{}
\InputIfFileExists{tables/panoptic2_numbers}{}{}
\newcommand{\xnum}[5]{\ifcsname #1@#2@#3@#4@#5\endcsname\csname #1@#2@#3@#4@#5\endcsname\else\missingnum{#1@#2@#3@#4@#5}\fi}
\newcommand{\xinfo}[3]{\ifcsname #1@#2@#3\endcsname\csname #1@#2@#3\endcsname\else\missingnum{#1@#2@#3}\fi}
\InputIfFileExists{tables/fit3d_numbers}{}{}
\InputIfFileExists{tables/panoptic_knownk_numbers}{}{}

\newcommand{\vx}{\mathbf{x}}
\newcommand{\vX}{\mathbf{X}}
\newcommand{\vd}{\mathbf{d}}
\newcommand{\ve}{\mathbf{e}}
\newcommand{\vr}{\mathbf{r}}
\newcommand{\vu}{\mathbf{u}}
\newcommand{\vy}{\mathbf{y}}
\newcommand{\vo}{\mathbf{o}}
\newcommand{\vt}{\mathbf{t}}
\newcommand{\mI}{\mathbf{I}}
\newcommand{\mP}{\mathbf{P}}
\newcommand{\mA}{\mathbf{A}}
\newcommand{\mK}{\mathbf{K}}
\newcommand{\mR}{\mathbf{R}}

%% file: tables/panoptic_numbers.tex
\expandafter\def\csname pn@2@subsets\endcsname{465}
\expandafter\def\csname pn@2@laf@picks\endcsname{$\lambda{=}0.01$, $\alpha{=}0.5$}
\expandafter\def\csname pn@2@ransac@mpjpe@worst\endcsname{284.3}
\expandafter\def\csname pn@2@laf@mpjpe@median\endcsname{29.4}
\expandafter\def\csname pn@2@laf@mpjpe@worst\endcsname{56.0}
\expandafter\def\csname pn@2@laf@mpjpe@beatsransac\endcsname{316}
\expandafter\def\csname pn@2@laf@pa@beatsransac\endcsname{443}
\expandafter\def\csname pn@2@dlt_lm@mpjpe@worst\endcsname{273.4}
\expandafter\def\csname pn@2@midpoint@mpjpe@worst\endcsname{88.2}
\expandafter\def\csname pn@3@subsets\endcsname{1{,}000}
\expandafter\def\csname pn@3@laf@picks\endcsname{$\lambda{=}0$, $\alpha{=}2$}
\expandafter\def\csname pn@3@dlt_irls@mpjpe@worst\endcsname{178.1}
\expandafter\def\csname pn@3@ransac@mpjpe@median\endcsname{24.9}
\expandafter\def\csname pn@3@ransac@mpjpe@worst\endcsname{61.8}
\expandafter\def\csname pn@3@laf@mpjpe@median\endcsname{25.2}
\expandafter\def\csname pn@3@laf@mpjpe@worst\endcsname{37.8}
\expandafter\def\csname pn@3@laf@mpjpe@p95\endcsname{31.9}
\expandafter\def\csname pn@3@laf@pa@beatsransac\endcsname{638}
\expandafter\def\csname pn@3@laf@mpjpe@rhoransac\endcsname{0.72}
\expandafter\def\csname pn@3@hybrid@mpjpe@p95\endcsname{30.7}
\expandafter\def\csname pn@3@hybrid@mpjpe@beatsransac\endcsname{753}
\expandafter\def\csname pn@3@hybrid@pa@beatsransac\endcsname{923}
\expandafter\def\csname pn@3@hybrid@mpjpe@rhoransac\endcsname{0.97}
\expandafter\def\csname pn@3@ransac_lm@mpjpe@median\endcsname{25.0}
\expandafter\def\csname pn@3@midpoint@mpjpe@worst\endcsname{39.8}
\expandafter\def\csname pn@4@subsets\endcsname{1{,}000}
\expandafter\def\csname pn@4@dlt_irls@mpjpe@worst\endcsname{467.0}
\expandafter\def\csname pn@4@ransac@mpjpe@median\endcsname{22.9}
\expandafter\def\csname pn@4@ransac@mpjpe@worst\endcsname{45.5}
\expandafter\def\csname pn@4@laf@mpjpe@median\endcsname{22.8}
\expandafter\def\csname pn@4@laf@mpjpe@worst\endcsname{32.3}
\expandafter\def\csname pn@4@laf@pa@beatsransac\endcsname{760}
\expandafter\def\csname pn@4@hybrid@mpjpe@beatsransac\endcsname{860}
\expandafter\def\csname pn@4@hybrid@pa@beatsransac\endcsname{975}
\expandafter\def\csname pn@4@ransac_lm@mpjpe@median\endcsname{23.0}
\expandafter\def\csname pn@4@midpoint@mpjpe@worst\endcsname{35.1}
\expandafter\def\csname pn@5@subsets\endcsname{50}
\expandafter\def\csname pn@20@ransac@mpjpe@median\endcsname{19.9}
\expandafter\def\csname pn@20@laf@mpjpe@median\endcsname{19.2}

%% file: tables/panoptic_depth.tex
\expandafter\def\csname db@framemin\endcsname{0.6}
\expandafter\def\csname db@framemax\endcsname{2.6}
\expandafter\def\csname db@framecor\endcsname{0.16}
\expandafter\def\csname db@frames\endcsname{2{,}534}
\expandafter\def\csname db@raw\endcsname{1.50}
\expandafter\def\csname db@cor\endcsname{59}
\expandafter\def\csname db@farther\endcsname{99.9}
\expandafter\def\csname db@pair\endcsname{1.8}

%% file: tables/panoptic_2d.tex
\expandafter\def\csname sd@median\endcsname{8.0}
\expandafter\def\csname sd@rel\endcsname{0.9}
\expandafter\def\csname sd@cammin\endcsname{6.3}
\expandafter\def\csname sd@cammax\endcsname{11.3}

%% file: tables/panoptic_pairs.tex
\expandafter\def\csname pg@narrowhi\endcsname{30}
\expandafter\def\csname pg@narrowpct\endcsname{14}
\expandafter\def\csname pg@top20narrow\endcsname{9}

%% file: tables/panoptic_ransac.tex
\expandafter\def\csname rt@taus\endcsname{10, 15, 20, 30, 50, 75, 100, 150, 200, 300}
\expandafter\def\csname rt@3@picks\endcsname{20}
\expandafter\def\csname rt@3@minworst\endcsname{50.9}
\expandafter\def\csname rt@3@heldout@median\endcsname{24.9}
\expandafter\def\csname rt@3@heldout@worst\endcsname{61.8}
\expandafter\def\csname rt@4@picks\endcsname{15}
\expandafter\def\csname rt@4@minworst\endcsname{43.3}
\expandafter\def\csname rt@4@heldout@median\endcsname{22.8}
\expandafter\def\csname rt@4@heldout@worst\endcsname{46.3}

%% file: tables/panoptic_failures.tex
\expandafter\def\csname fa@gross\endcsname{500}
\expandafter\def\csname fa@2@ransac@pct\endcsname{0.23}
\expandafter\def\csname fa@2@ransac@max\endcsname{836.6}
\expandafter\def\csname fa@2@laf@pct\endcsname{0.20}
\expandafter\def\csname fa@2@laf@max\endcsname{5.3}
\expandafter\def\csname fa@3@dlt_irls@max\endcsname{1040.7}
\expandafter\def\csname fa@3@ransac@pct\endcsname{0.08}
\expandafter\def\csname fa@3@ransac@max\endcsname{122.4}
\expandafter\def\csname fa@3@laf@pct\endcsname{0.13}
\expandafter\def\csname fa@3@laf@max\endcsname{3.4}

%% file: tables/panoptic_gold.tex
\expandafter\def\csname gs@2@midpoint@max\endcsname{109.4}

%% file: tables/panoptic_ablation_numbers.tex
\expandafter\def\csname ab@3@subsets\endcsname{1{,}000}
\expandafter\def\csname ab@3@mean_raw@mpjpe\endcsname{873.3}
\expandafter\def\csname ab@3@mean@mpjpe\endcsname{56.9}
\expandafter\def\csname ab@3@uniform@mpjpe\endcsname{26.5}
\expandafter\def\csname ab@3@full_residual@mpjpe\endcsname{26.1}
\expandafter\def\csname ab@3@laf@mpjpe\endcsname{25.2}
\expandafter\def\csname ab@3@isotropic@mpjpe\endcsname{63.4}
\expandafter\def\csname ab@5@subsets\endcsname{50}
\expandafter\def\csname ab@5@full_residual@mpjpe\endcsname{23.9}
\expandafter\def\csname ab@5@laf@mpjpe\endcsname{22.1}

%% file: tables/panoptic_runtime.tex
\expandafter\def\csname tm@cpu\endcsname{12th Gen Intel Core i9-12900H}
\expandafter\def\csname tm@points\endcsname{18{,}509}
\expandafter\def\csname tm@4@laf\endcsname{53}
\expandafter\def\csname tm@4@ransac\endcsname{416}
\expandafter\def\csname tm@4@dlt_irls@ratio\endcsname{4.3}
\expandafter\def\csname tm@31@laf\endcsname{331}
\expandafter\def\csname tm@31@ransac\endcsname{54{,}356}
\expandafter\def\csname tm@31@dlt_irls@ratio\endcsname{1.6}

%% file: tables/panoptic_inference.tex
\expandafter\def\csname ti@sperframe\endcsname{0.25}
\expandafter\def\csname ti@gpu\endcsname{GeForce RTX 3070 Ti Laptop GPU}

%% file: tables/panoptic_correction_numbers.tex
\expandafter\def\csname dc@raw@mpjpe\endcsname{1{,}079}
\expandafter\def\csname dc@single@pagain@max\endcsname{45}
\expandafter\def\csname dc@single@pagain@median\endcsname{13.3}
\expandafter\def\csname dc@single@shift@mpjpe\endcsname{78.7}
\expandafter\def\csname dc@mean@pagain\endcsname{9.1}
\expandafter\def\csname dc@pairs\endcsname{465}
\expandafter\def\csname dc@known@single@mpjpe\endcsname{85.6}
\expandafter\def\csname dc@known@single@better\endcsname{28}

%% file: tables/panoptic_bones_numbers.tex
\expandafter\def\csname bl@thigh@min\endcsname{32.9}
\expandafter\def\csname bl@thigh@max\endcsname{43.9}
\expandafter\def\csname bl@spread\endcsname{33}

%% file: tables/panoptic_supp_numbers.tex
\expandafter\def\csname sj@subsets\endcsname{1{,}000}

%% file: tables/panoptic_estimators.tex
\expandafter\def\csname es@sam@ratio@min\endcsname{1.18}
\expandafter\def\csname es@sam@ratio@max\endcsname{1.58}
\expandafter\def\csname es@sam@single@mpjpe\endcsname{78.7}
\expandafter\def\csname es@hm@single@mpjpe\endcsname{189.6}
\expandafter\def\csname es@cf@single@mpjpe\endcsname{120.6}

%% file: tables/qualitative_numbers.tex
\expandafter\def\csname ql@thr\endcsname{25}
\expandafter\def\csname ql@a@dlt@jointm\endcsname{17}
\expandafter\def\csname ql@a@midpoint@joint\endcsname{left eye}
\expandafter\def\csname ql@a@midpoint@jointm\endcsname{16}
\expandafter\def\csname ql@a@midpoint@jointangle\endcsname{0.1}
\expandafter\def\csname ql@a@midpoint@jointpx\endcsname{14}
\expandafter\def\csname ql@b@badview\endcsname{00\_15}
\expandafter\def\csname ql@b@badnames\endcsname{left elbow and left wrist}
\expandafter\def\csname ql@b@badpx@min\endcsname{77}
\expandafter\def\csname ql@b@badpx@max\endcsname{282}
\expandafter\def\csname ql@b@badweight\endcsname{0.09}
\expandafter\def\csname ql@c@ransac@joint\endcsname{right elbow}
\expandafter\def\csname ql@c@ransac@jointangle\endcsname{0.6}
\expandafter\def\csname ql@c@ransac@inliers\endcsname{00\_16 and 00\_26}

%% file: tables/panoptic_detector.tex
\expandafter\def\csname dt@px\endcsname{9.4}
\expandafter\def\csname dt@missing\endcsname{20.9}
\expandafter\def\csname dt@missing@ankle\endcsname{41}
\expandafter\def\csname dt@outside@ankle\endcsname{44}
\expandafter\def\csname dt@2@coverage\endcsname{66}
\expandafter\def\csname dt@2@best@mpjpe@median\endcsname{37.8}
\expandafter\def\csname dt@2@best@mpjpe@worst\endcsname{1{,}213.9}
\expandafter\def\csname dt@3@best@mpjpe@median\endcsname{30.4}
\expandafter\def\csname dt@3@best@mpjpe@worst\endcsname{137.4}
\expandafter\def\csname dt@4@coverage\endcsname{91}
\expandafter\def\csname dt@4@best@mpjpe@median\endcsname{26.8}
\expandafter\def\csname dt@4@best@mpjpe@worst\endcsname{46.6}

%% file: tables/panoptic2_numbers.tex
\expandafter\def\csname p2@name\endcsname{171026\_pose1}
\expandafter\def\csname p2@scored\endcsname{425}
\expandafter\def\csname p2@subjects\endcsname{3}

%% file: tables/fit3d_numbers.tex
\expandafter\def\csname fr@subjects\endcsname{8}
\expandafter\def\csname fr@frames\endcsname{17{,}970}
\expandafter\def\csname fr@dist@min\endcsname{4.3}
\expandafter\def\csname fr@dist@max\endcsname{4.5}
\expandafter\def\csname fr@bad@cross\endcsname{17}
\expandafter\def\csname fr@bad@dlt@min\endcsname{81.9}
\expandafter\def\csname fr@bad@dlt@max\endcsname{95.8}
\expandafter\def\csname fr@bad@laf@min\endcsname{63.8}
\expandafter\def\csname fr@bad@laf@max\endcsname{65.4}
\expandafter\def\csname fr@good@lafgap@max\endcsname{2.3}
\expandafter\def\csname fr@2@dlt_irls@mpjpe@worst\endcsname{95.8}
\expandafter\def\csname fr@2@midpoint@mpjpe@worst\endcsname{77.7}
\expandafter\def\csname fr@2@laf@mpjpe@worst\endcsname{65.4}
\expandafter\def\csname fr@3@rejected\endcsname{0.3}
\expandafter\def\csname fr@34@spread\endcsname{0.7}

%% file: tables/panoptic_knownk_numbers.tex
\expandafter\def\csname kk@2@dlt_irls@mpjpe@worst\endcsname{220.8}
\expandafter\def\csname kk@3@dlt_irls@mpjpe@worst\endcsname{820.2}

%% file: sec/0_abstract.tex
\begin{abstract}
Put a monocular mesh estimator behind every camera of a calibrated rig, and the reconstructed bodies land a meter or more apart.
The cause is geometric: each estimate sits at the depth implied by a focal length the network assumed, not the one the camera has.
Its error therefore points almost entirely along its own viewing ray, and the calibration tells us that direction exactly.
To address this, we introduce the Ray-Lateral Agreement (\method), which follows one rule: \emph{trust the ray, not the depth}.
\method first slides every estimate along its own rays, which removes the depth bias and leaves the shape untouched.
It then fuses the views one point at a time, under a Gaussian error whose precision follows each viewing ray.
The maximum-likelihood solution is closed-form, interpolates between averaging and midpoint triangulation, and stays well conditioned where triangulation is singular.
Robustness is the harder half, because correct views also disagree along their rays.
Any weight built on the full residual therefore punishes good views together with bad ones.
Ray-lateral agreement instead scores every view, at every point, by its residual across that view's own ray alone.
We evaluated \method on multi-camera public benchmark datasets. Across two, three, and four cameras and three per-view pose estimators, \method consistently achieves lower worst-case error than all triangulation baselines.
\end{abstract}

%% file: sec/1_intro.tex
\section{Introduction}
\label{sec:intro}

Monocular human mesh recovery has advanced remarkably. Given a single image, models such as HMR~\cite{kanazawa2018hmr}, HMR~2.0~\cite{goel2023humans}, and SAM~3D~Body~\cite{yang2026sam3dbody} can reconstruct a complete, plausible 3D human body. At the same time, applications in gait assessment, sports science, and rehabilitation are increasingly shifting from marker-based motion-capture studios to setups with a few calibrated consumer cameras~\cite{uhlrich2023opencap,kanko2021concurrent,pagnon2021pose2sim}. One might therefore expect that applying these models independently to each camera view would yield consistent 3D reconstructions. Surprisingly, the resulting bodies can disagree in position by more than a meter.
The reason is geometric. A monocular network converts apparent size into metric depth through a focal length it has to assume, usually a fixed default~\cite{kissos2020beyond}, which real cameras rarely match: each per-view body lands at the wrong distance along its own camera's rays, by an amount that differs from camera to camera (\cref{fig:teaser}a).
On CMU Panoptic~\cite{joo2015panoptic}, \gnum{db}{farther}\% of the views place the person farther from their camera than the reference, each by a different amount (\cref{fig:disagreement}).

Neither standard remedy resolves this.
Averaging the bodies inherits the bias.
Triangulating their 2D keypoints~\cite{hartley1997triangulation,hartley2004multiple}, robustly with RANSAC~\cite{fischler1981ransac} or reweighting~\cite{holland1977robust}, discards the 3D information each network provides, and is most fragile exactly where small rigs operate: with two cameras no test can find a bad view, and nearly collinear rays make the solution degenerate.
Learned multi-view fusion~\cite{iskakov2019learnable,qiu2019cross,he2020epipolar,tu2020voxelpose} needs training data for the target rig and transfers poorly to new camera layouts~\cite{bartol2022generalizable}.

Our key observation is that the error of a monocular estimate has a known direction.
Reprojected through its own assumed camera, the estimate fits the image closely: it is accurate across its viewing ray and inaccurate along it, and it stays so after the depth correction (\cref{sec:setup,sec:exp_panoptic}).
In a calibrated rig, that ray is known for every view and every point.
We turn this into a principle, \emph{trust the ray, not the depth}, and into a fusion operator, \method (\cref{fig:overview}), that
\begin{enumerate*}[label=(\roman*)]
\item corrects the depth by translating each estimate along its rays, which keeps every point on its observed ray and the body's extent intact;
\item fuses the views point by point with an anisotropic precision that trusts each camera fully across its ray and only by a factor $\lambda\ll1$ along it; and
\item weights each camera at each point by its agreement with the consensus \emph{across} its ray alone.
\end{enumerate*}
A depth-ambiguous estimator disagrees along its own ray even when it is right; disagreement across the ray is the evidence that a view is wrong.
\method is closed-form and training-free, and costs three $3\times3$ solves per point.
It applies to any per-view estimator whose outputs correspond across views, and we test it with three: SAM~3D~Body, HMR~2.0 and CameraHMR.
For an estimator that is given the true intrinsics, step~(i) is the identity.

Our contributions are:
\begin{itemize}
\item an analysis of the focal-induced depth bias: the natural fix, rescaling the depth, compresses the body along its ray and degrades every single view, whereas translating along the ray does not;
\item \method, a training-free operator with a maximum-likelihood interpretation, which combines an anisotropic per-view precision with a scale-free, ray-lateral agreement weight for every camera and every point; and
\item a camera-placement study against DLT, RANSAC, midpoint and reprojection-error triangulation, with the same inputs, camera subsets and held-out parameters for every method: thousands of camera subsets of two CMU Panoptic sequences, with three per-view estimators, and every camera subset of the Fit3D fitness dataset.
\end{itemize}
Our experiments show that \method degrades gracefully where triangulation breaks down.
With two to four cameras, its worst case is below that of every triangulation baseline on both Panoptic sequences, and its parameters transfer unchanged to the second sequence and to Fit3D, where it avoids the failures of triangulation between cameras that face each other.
Ray-lateral agreement also complements RANSAC's view selection: fused within RANSAC's consensus set, it has the lowest median error on Panoptic with SAM~3D~Body at every camera count we tested.

%% file: sec/2_related.tex
\section{Related work}
\label{sec:related}

\begin{figure*}[t]
  \centering
  \includegraphics[width=\linewidth]{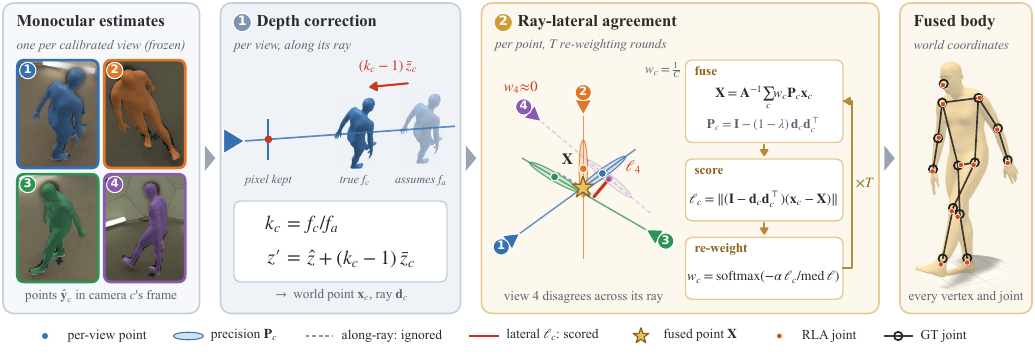}
  \caption{\textbf{Overview of \method.}
  A frozen monocular estimator reconstructs the person independently in each calibrated view.
  (1)~Depth correction (\cref{sec:depth}) slides each body along its rays by $(k_c-1)\bar z_c$, where $k_c=f_c/f_a$ is the ratio of the true to the assumed focal length, keeping every point on the ray through the pixel the estimator committed to.
  (2)~Ray-lateral agreement (\cref{sec:aniso,sec:agree}) fuses every point separately: it trusts each view across its ray but hardly along it, and weights it by its lateral residual to the current fused point alone; fusing, scoring and re-weighting repeat $T$ times.
  A view that disagrees across its ray (4) gets almost no weight, while views that differ only in depth (1--3) agree.
  The output is a single body in world coordinates; its joints closely match the ground truth.}
  \label{fig:overview}
\end{figure*}

\paragraph{Multi-view 3D human pose.}
Calibrated multi-view capture classically triangulates 2D keypoints~\cite{hartley1997triangulation,hartley2004multiple} or fits a body model to the keypoints and silhouettes of all views~\cite{huang2017towards}.
Learned methods replace these steps with trained components: algebraic and volumetric triangulation~\cite{iskakov2019learnable}, cross-view feature fusion~\cite{qiu2019cross}, epipolar attention~\cite{he2020epipolar}, camera-disentangled latent codes~\cite{remelli2020lightweight}, and volumetric~\cite{tu2020voxelpose}, direct-regression~\cite{wang2021mvp} or cross-view matching~\cite{dong2019fast} models for several people.
They learn their fusion for the camera layouts they were trained on, and degrade on new ones~\cite{bartol2022generalizable}.
In markerless biomechanics, OpenCap~\cite{uhlrich2023opencap} and Pose2Sim~\cite{pagnon2021pose2sim} triangulate multi-view 2D keypoints to reconstruct 3D motion for musculoskeletal analysis, while studies of commercial markerless systems have evaluated their accuracy against marker-based motion capture~\cite{kanko2021concurrent}. All of these combine 2D evidence across views. We instead combine the complete 3D bodies that a monocular network now predicts in every view, without training for the rig, and study how the result depends on the number and placement of the cameras.
\vspace{-2mm}
\paragraph{Monocular mesh recovery and its depth.}
Monocular methods regress body-model parameters and a camera from a single crop~\cite{kanazawa2018hmr,kolotouros2019spin,goel2023humans,yang2026sam3dbody}.
The true intrinsics are usually unknown, so they assume a weak-perspective or fixed-focal camera, which ties the estimated depth to that assumption~\cite{kissos2020beyond}.
CLIFF~\cite{li2022cliff} feeds the crop's location in the full frame to the network, SPEC~\cite{kocabas2021spec} estimates the camera, and SAM~3D~Body~\cite{yang2026sam3dbody} can be conditioned on given intrinsics.
These advances improve each single view, yet depth remains its least certain part, because it follows from the apparent size of a body whose true size is unknown.
We take the per-view estimates as they are, remove their depth bias with the known calibration, and design the fusion around the ambiguity that remains.
\vspace{-2mm}
\paragraph{Robust and uncertainty-aware fusion.}
RANSAC~\cite{fischler1981ransac} and iteratively reweighted least squares with a robust loss~\cite{huber1964robust,holland1977robust} are the classical answers to outlier views.
Weighting each observation by its covariance is standard in photogrammetry and bundle adjustment~\cite{forstner2016geometry,triggs2000bundle}.
Depth-sensor noise models likewise separate the error along the viewing ray from the error across it~\cite{nguyen2012modeling}.
Learned multi-view methods predict one scalar confidence per view~\cite{iskakov2019learnable,zhang2021adafuse}.
MV-SAM3D~\cite{li2026mv} fuses views of objects in the latent space of a 3D generative model, and weights each view by attention entropy and visibility.
A scalar confidence cannot express that a view is reliable perpendicular to its viewing ray but uncertain along it. A fixed anisotropic noise model, in turn, cannot distinguish a consistent view from an outlier. \method addresses both in closed form. It assigns low precision along each viewing ray and high precision perpendicular to it, and computes a weight for every camera and every point from cross-view agreement measured perpendicular to the camera's ray. A view is therefore not penalized for its error along its own ray, which is precisely the depth error that dominates monocular estimates.

%% file: sec/3_method.tex
\vspace{-3.8mm}
\section{Method}
\label{sec:method}

We assume $C$ synchronised, calibrated cameras.
For each camera $c$ and frame, a monocular 3D estimator returns $P$ points $\hat\vy_{c,p}\in\mathbb{R}^3$ in that camera's coordinate frame.
Our main estimator is SAM~3D~Body~\cite{yang2026sam3dbody}, which returns $18{,}439$ mesh vertices and $70$ keypoints per view.
Any estimator whose outputs correspond across views would do, because the point index gives the correspondence; we match no features.
Camera $c$ has intrinsics $\mK_c$ with focal length $f_c$ (the mean of its two focal lengths), a lens-distortion model, and extrinsics $(\mR_c,\vt_c)$ that map world to camera coordinates.
Its centre is $\vo_c=-\mR_c^\top\vt_c$.
We process every frame, and every point within it, independently, and drop the indices where they are unambiguous.
Sections~\ref*{sec:supp_mle}--\ref*{sec:supp_seq2} of the supplementary material give derivations, baselines, details and further results.

\subsection{Correcting the depth bias}
\label{sec:depth}

\begin{figure}[t]
  \centering
  \includegraphics[width=0.92\linewidth]{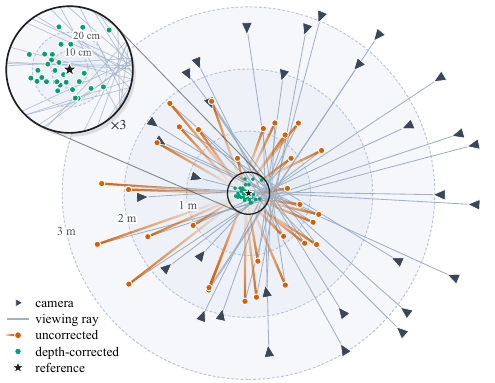}
  \caption{\textbf{The depth bias on CMU Panoptic} (plan view, one frame).
  Each of the 31 HD cameras (triangles) reconstructs the person independently; we place each pelvis in world coordinates with the known calibration.
  Uncorrected (orange, joined to its camera), the estimates lie \gnum{db}{framemin}--\gnum{db}{framemax}\,m from the reference; rings mark the distance in the floor plane.
  The correction of \cref{eq:shift} moves each one back towards its camera (fading trail), to a median error of \gnum{db}{framecor}\,m (green; lens $\times3$).
  Over all \gnum{db}{frames} scored frames, \gnum{db}{farther}\% of uncorrected views place the pelvis farther from their camera than the reference; the median error falls from \gnum{db}{raw}\,m to \gnum{db}{cor}\,mm.}
  \label{fig:disagreement}
\end{figure}

The estimator assumes a pinhole camera with focal length $f_a$ and principal point at the image centre.
Under that assumption it places its reconstruction at the depth that explains the observed image size (\cref{fig:disagreement}).
Reprojecting the estimate through its own assumed intrinsics,
\begin{equation}
  \vu_{c,p}=\frac{f_a}{\hat y_{z}}\begin{pmatrix}\hat y_{x}\\\hat y_{y}\end{pmatrix}+\frac12\begin{pmatrix}W\\H\end{pmatrix},
  \label{eq:reproj}
\end{equation}
recovers the pixel the network committed to, which is accurate even though the depth is not.
We undistort $\vu_{c,p}$ with the true lens model and back-project it through $\mK_c$, which gives the exact viewing ray $\vr_{c,p}=\mK_c^{-1}(\bar\vu_{c,p},1)^\top$, scaled to unit depth.

\begin{figure*}[t]
  \centering
  \includegraphics[width=0.92\linewidth]{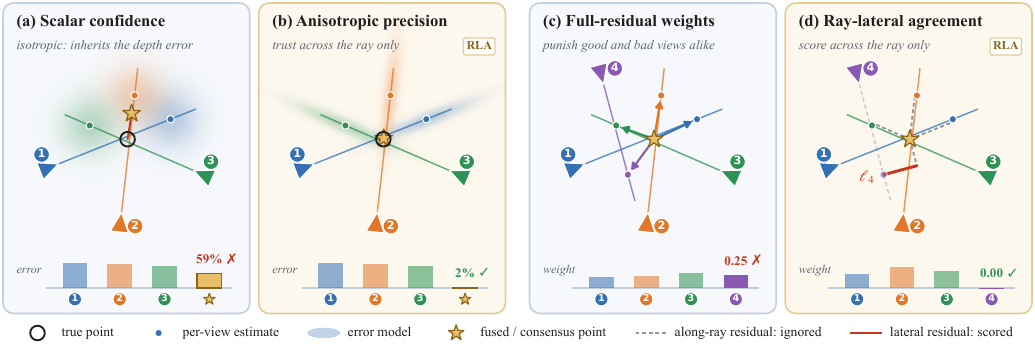}
  \caption{\textbf{The two ingredients of \method, each next to the alternative it replaces} (2D; fused points and weights computed with \cref{eq:fuse,eq:weights}).
  Each per-view estimate (dot) lies on its own camera's (triangle) viewing ray, but too far along it; the three views share this depth error.
  (a)~A scalar confidence gives every view an isotropic error (shading), so the fused point (star) keeps most of that shared error; (b)~the anisotropic precision of \cref{eq:proj} ($\lambda=0.02$) trusts each view across its ray only, and the fused point falls on the true one (bars: the error of each view and of the fused point).
  (c)~A fourth view is wrong across its ray; weighted by its full residual to the consensus point, it keeps as much weight (bars) as the correct views, whose residuals are long but lie along their rays.
  (d)~Ray-lateral agreement (\cref{eq:lateral,eq:weights}, $\alpha=2$) scores only the part across each ray (red) and ignores the part along it (grey), so the wrong view gets no weight.}
  \label{fig:method}
\end{figure*}

The depth must then change by the factor $k_c=f_c/f_a$: under the true focal length, an object of fixed metric size that looks equally large lies $k_c$ times as far away.
The obvious correction multiplies every depth by $k_c$, but that also scales the body's own extent along the ray by $k_c$, which is $0.63$--$0.84$ on Panoptic.
Instead we translate the estimate along the ray, which preserves each point's depth offset from the body:
\begin{equation}
  z'_{c,p}=\hat y_{z,c,p}+(k_c-1)\,\bar z_c,\qquad
  \bar z_c=\median_{j\in\mathcal J}\hat y_{z,c,j},
  \label{eq:shift}
\end{equation}
where $\mathcal J$ are the body joints.
The corrected point $\vy'_{c,p}=z'_{c,p}\,\vr_{c,p}$ lies exactly on the observed ray, and in world coordinates
\begin{equation}
  \vx_{c,p}=\mR_c^\top(\vy'_{c,p}-\vt_c),\qquad
  \vd_{c,p}=\frac{\vx_{c,p}-\vo_c}{\lVert\vx_{c,p}-\vo_c\rVert}.
  \label{eq:world}
\end{equation}
\Cref{eq:shift} keeps every point on its observed ray and keeps its depth offset exactly.
Lateral offsets change only by $(k_c^{-1}-1)\,a\,\delta$, where $a$ is the point's normalised image coordinate and $\delta$ its depth offset from $\bar z_c$ (\cref*{sec:supp_shift}).
Rescaling, by contrast, compresses every depth offset by the factor $k_c$.
In \cref*{tab:depth} (supplement), rescaling costs a single view up to \gnum{dc}{single@pagain@max}\,mm PA-MPJPE.

\subsection{Anisotropic fusion}
\label{sec:aniso}

A monocular estimate is precise across its viewing ray and imprecise along it (\cref{fig:method}a,b).
We encode this with the precision
\begin{equation}
  \mP_{c,p}=(\mI-\vd_{c,p}\vd_{c,p}^\top)+\lambda\,\vd_{c,p}\vd_{c,p}^\top,\qquad0\le\lambda\le1,
  \label{eq:proj}
\end{equation}
and fuse point $p$ as the weighted least-squares solution
\begin{equation}
\begin{aligned}
  \vX_p&=\argmin_{\vX}\sum_c w_{c,p}(\vX-\vx_{c,p})^\top\mP_{c,p}(\vX-\vx_{c,p})\\
  &=\mA_p^{-1}\sum_c w_{c,p}\mP_{c,p}\vx_{c,p},
\end{aligned}
  \label{eq:fuse}
\end{equation}
with $\mA_p=\sum_c w_{c,p}\mP_{c,p}$ and non-negative weights $w_{c,p}$ that sum to one over cameras.
\Cref{eq:fuse} is the maximum-likelihood estimate when each view's error is Gaussian with covariance proportional to $(\mI-\vd\vd^\top)+\lambda^{-1}\vd\vd^\top$: the along-ray variance is $1/\lambda$ times the lateral one (\cref*{sec:supp_mle}).
It interpolates between two familiar estimators: for $\lambda=1$ the weighted mean of the per-view points, and as $\lambda\to0$ least-squares (midpoint) triangulation~\cite{hartley1997triangulation} of the rays through the observed pixels.
For $\lambda>0$ every eigenvalue of $\mA_p$ is at least $\lambda$, so the solve stays well conditioned when the rays are nearly parallel, where triangulation is singular; a single camera returns its own estimate.

\subsection{Ray-lateral agreement}
\label{sec:agree}
The weights should say how far to trust each view at each point; we measure this from cross-view agreement.
Given a current fused estimate $\vX_p$, we keep only the part of each view's residual $\ve_{c,p}=\vx_{c,p}-\vX_p$ that is perpendicular to the view's own ray,
\begin{equation}
  \ell_{c,p}=\bigl\lVert(\mI-\vd_{c,p}\vd_{c,p}^\top)\,\ve_{c,p}\bigr\rVert .
  \label{eq:lateral}
\end{equation}
A depth-ambiguous estimator produces the along-ray part even when it is right, so we discard it (\cref{fig:method}c,d): only the lateral residual, measured in the direction in which the view is precise, is evidence that the view is wrong.
We normalise it by its median over cameras and turn it into weights with a softmax:
\begin{equation}
  s_{c,p}=-\frac{\ell_{c,p}}{\median_{c'}\ell_{c',p}+\epsilon},\qquad
  w_{c,p}=\frac{\exp(\alpha s_{c,p})}{\sum_{c'}\exp(\alpha s_{c',p})}.
  \label{eq:weights}
\end{equation}
The median makes the score dimensionless, so a single sharpness $\alpha$ serves every point, whatever the body size, distance or rig scale, and it is robust to the very views it scores.
The weight is per camera and per point: we may trust one view at the shoulder and distrust the same view at an occluded wrist in the same frame.

We start from uniform weights, fuse with \cref{eq:fuse}, recompute the weights with \cref{eq:weights}, and fuse again, for $T=2$ re-weighting rounds; each round sharpens the weights, and a third or fourth round raises the median error slightly (\cref*{sec:supp_baselines}).
This is iteratively reweighted least squares~\cite{holland1977robust} with an anisotropic quadratic form and an exponential weight on the normalised lateral residual.
In total it costs $T+1=3$ solves of a $3\times3$ symmetric positive-definite system per point, with no training, no gradient-based optimisation and no estimator uncertainty.

%% file: sec/4_experiments.tex
\section{Experiments}
\label{sec:experiments}

\subsection{Setup}
\label{sec:setup}

\paragraph{CMU Panoptic.}
We use sequence \texttt{171204\_pose1} of the Panoptic Studio~\cite{joo2015panoptic,joo2017panoptic}: 31 synchronised HD cameras ($1920\times1080$, 29.97\,fps, layout: \cref*{fig:supp_cameras} in \cref*{sec:supp_panoptic}) with 3D reference keypoints reconstructed by the studio.
Six different people perform range-of-motion movements one after another, about 2.5 minutes each, with one person in view throughout; their thigh lengths range from \gnum{bl}{thigh@min} to \gnum{bl}{thigh@max}\,cm, so the six reference identities are different people (\cref*{tab:supp_bones,fig:supp_subjects} in \cref*{sec:supp_panoptic}).
We process every 10th frame, each with a single reference body.
We keep the frames in which all 17 COCO joints have reference confidence above $0.1$ (\gnum{db}{frames} frames), and map the Panoptic studio's 19 joints to the 17 COCO joints (\cref*{sec:supp_panoptic}).
Studies over many camera subsets use every 5th of these frames (507).
\vspace{-2mm}
\paragraph{Per-view estimates.}
We process each camera view independently. YOLOv8x~\cite{jocher2023yolov8} detects the person with a 12\% bounding-box margin, and SAM~3D~Body~\cite{yang2026sam3dbody} (ViT-H) reconstructs the body using its default focal length, $f_a=\sqrt{W^2+H^2}\approx2203$\,px. This assumed focal length is \gnum{es}{sam@ratio@min}--\gnum{es}{sam@ratio@max}$\times$ the true focal length of the Panoptic cameras. Notably, Panoptic was not included in the training data of SAM~3D~Body~\cite{yang2026sam3dbody}. Across 31 cameras and \gnum{db}{frames} frames, the reconstructed keypoints achieve a median 2D reprojection error of \gnum{sd}{median}\,px (\gnum{sd}{cammin}--\gnum{sd}{cammax}\,px across cameras), corresponding to just \gnum{sd}{rel}\% of the person's bounding-box diagonal. Thus, the reconstructions align well with the image evidence despite substantial errors in depth.
\vspace{-3mm}
\paragraph{Baselines.}
All methods receive identical inputs: the same per-view estimates, corrected rays and calibration.
\emph{Mean} and \emph{median} combine the corrected per-view 3D keypoints.
The triangulation baselines use the per-view keypoints reprojected with \cref{eq:reproj} and undistorted.
The direct linear transform (DLT)~\cite{hartley2004multiple} solves the stacked linear projection equations of all views by SVD, minimising an algebraic error rather than a geometric one.
\emph{DLT+IRLS} refines DLT by iteratively reweighted least squares (IRLS) with binary weights: two rounds reject the views whose reprojection error exceeds $\max(3\times\text{median},20\,\text{px})$, reverting to all views whenever fewer than four would remain, so with four or fewer cameras it is plain DLT; \emph{IRLS-2} may reject down to two views.
\emph{RANSAC}~\cite{fischler1981ransac} tries every camera pair as a hypothesis, keeps the largest consensus set within 20\,px, breaking ties by total reprojection error, and refits with DLT.
\emph{DLT+LM} and \emph{RANSAC+LM} refine these solutions by Levenberg--Marquardt minimisation of the reprojection error, over all views and over the consensus set respectively: the maximum-likelihood estimate under Gaussian image noise~\cite{hartley2004multiple}.
\emph{Midpoint} triangulation is the point closest to the viewing rays in least squares, \ie \cref{eq:fuse} with $\lambda=0$ and equal weights; \emph{uniform anisotropic} is \cref{eq:fuse} with equal weights; \emph{RANSAC set + \method}, the hybrid, applies \method within RANSAC's consensus set at each point.
We also run the standard 2D pipeline: a 2D pose detector, YOLOv8x-pose~\cite{jocher2023yolov8}, on the same frames and boxes, triangulated by DLT, by confidence-weighted DLT~\cite{iskakov2019learnable} and by RANSAC (\cref*{sec:supp_baselines}).
\vspace{-2mm}
\paragraph{Metrics.}
We report Mean Per-Joint Position Error (MPJPE) and Procrustes-Aligned MPJPE (PA-MPJPE), both in millimeters, over the 17 COCO joints.For experiments involving camera subsets, we report the median and maximum error across subsets (the 95th percentile in \cref*{fig:camera_count} of the supplement), where each subset's error is averaged over subjects. We also count the subsets in which one method outperforms another. MPJPE is the more stringent metric for our task, as it captures the residual depth errors that fusion aims to correct, along with discrepancies between the joint definitions used by the estimator and the ground truth.
\vspace{-2 mm}
\paragraph{Parameters.}
On Panoptic, we select $(\lambda,\alpha)$ from the grid $\{0,0.005,0.01,0.02\}\times\{0.5,1,2,4\}$ using leave-one-subject-out cross-validation.
For each held-out subject, we use the parameters that minimize MPJPE on the other five subjects, ensuring that no reported result is tuned on its evaluation subject.
\method processes all $18{,}439$ mesh vertices and $70$ keypoints per frame as a batch of independent points, with $\epsilon=10^{-6}$\,m in \cref{eq:weights}.

\subsection{The depth bias and its correction}
\label{sec:exp_depth}

All methods fuse the same depth-corrected estimates (\cref{sec:depth}), so we first test this shared step: how large the bias is, why we translate rather than rescale, and whether the comparison depends on the correction (full results in \cref*{tab:depth} in the supplement).
Without correction the per-view bodies are far apart: two views place the pelvis \gnum{db}{pair}\,m apart on average, and averaging two views gives \gnum{dc}{raw@mpjpe}\,mm MPJPE (median over pairs).
Both corrections remove the bias, but rescaling the depth by $k_c$ also compresses the body along the ray: it raises the PA-MPJPE of every single view, by a median of \gnum{dc}{single@pagain@median}\,mm and up to \gnum{dc}{single@pagain@max}\,mm, and makes the mean of two views \gnum{dc}{mean@pagain}\,mm worse.
\method is nearly insensitive to the choice, because it trusts each view's depth only through $\lambda$; we use translation throughout.

SAM~3D~Body also accepts each camera's true intrinsics as input, and then needs no correction (\cref*{tab:depth}, right).
This does not place the single views better: our correction of the default output has the lower MPJPE, \gnum{dc}{single@shift@mpjpe} against \gnum{dc}{known@single@mpjpe}\,mm, on \gnum{dc}{known@single@better} of the 31 cameras, with a similar PA-MPJPE, and after fusion the two inputs differ by less than a millimetre.
With the true intrinsics, DLT still fails badly on a few subsets (worst \xnum{kk}{2}{dlt_irls}{mpjpe}{worst}\,mm with two cameras, \xnum{kk}{3}{dlt_irls}{mpjpe}{worst}\,mm with three), so the triangulation failures in \cref{sec:exp_panoptic} do not come from the assumed focal length.

\subsection{Camera placement and count on Panoptic}
\label{sec:exp_panoptic}

\begin{table*}[t]
  \centering
  \small
  \setlength{\tabcolsep}{4.5pt}
  \renewcommand{\arraystretch}{0.95}
  \input{tables/combined_main}
  \caption{\textbf{Camera subsets of CMU Panoptic and Fit3D} (mm; median and worst over subsets, random ones where not all). Panoptic: 507 frames, six subjects, 17 joints; Fit3D: \gnum{fr}{frames} frames, \gnum{fr}{subjects} subjects, 12 limb joints.
  With two cameras no baseline can reject a view, so DLT+IRLS, IRLS-2 and RANSAC reduce to DLT, RANSAC+LM to DLT+LM, and the hybrid to \method.
  Held-out: $(\lambda,\alpha)$ by leave-one-subject-out on Panoptic, reused on Fit3D. Bold: best per column and dataset.}
  \label{tab:panoptic}
\end{table*}

\begin{figure*}[t]
  \centering
  \includegraphics[width=0.9\linewidth]{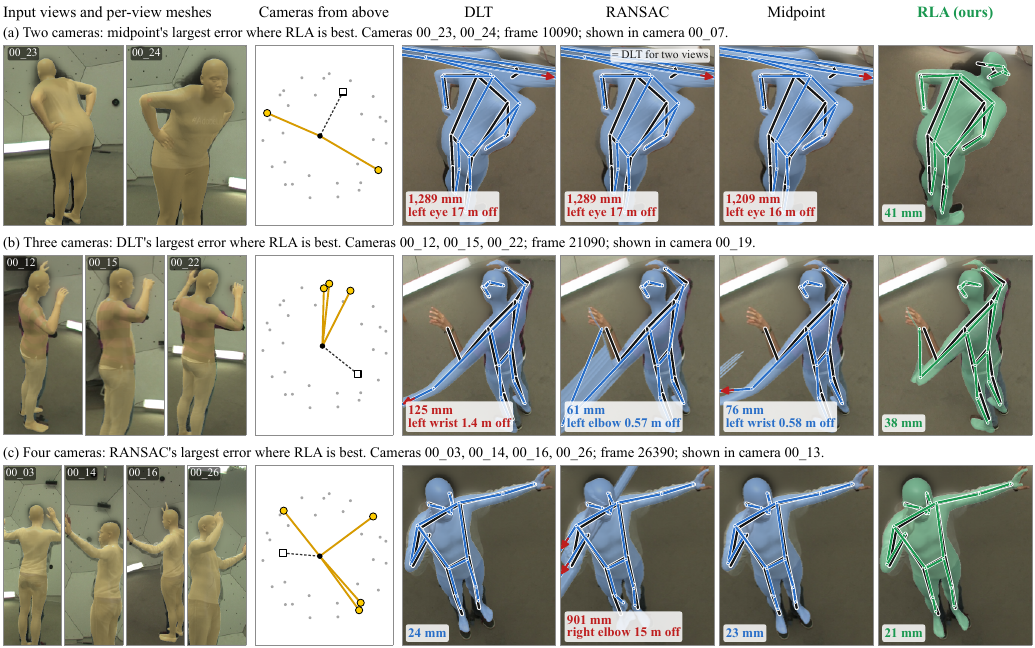}
  \caption{\textbf{Qualitative results} on Panoptic, chosen by rule: among the subsets and frames of \cref{tab:panoptic} where the keypoints of every input view are mostly accurate (median error below \gnum{ql}{thr}\,px) and \method is the most accurate method, the largest error of (a) midpoint triangulation with two cameras, (b) DLT with three, (c) RANSAC with four (\cref*{sec:supp_qualitative}).
  Each method fuses every vertex of the per-view meshes (left) as it fuses the joints; its mesh and skeleton (colour) are drawn with the reference skeleton (black) in a camera that no method used (square), with the MPJPE of its joints; arrows mark joints that leave the panel.
  In (a), the rays of the \gnum{ql}{a@midpoint@joint} are \gnum{ql}{a@midpoint@jointangle}$^\circ$ apart, so a \gnum{ql}{a@midpoint@jointpx}\,px keypoint error places it \gnum{ql}{a@midpoint@jointm}\,m away.
  In (b), camera \gnum{ql}{b@badview} misplaces the \gnum{ql}{b@badnames} by \gnum{ql}{b@badpx@min}--\gnum{ql}{b@badpx@max}\,px; DLT and midpoint follow it, RANSAC rejects only the wrist, and \method weights them at most \gnum{ql}{b@badweight}.
  In (c), RANSAC triangulates the \gnum{ql}{c@ransac@joint} from cameras \gnum{ql}{c@ransac@inliers} alone, whose rays are \gnum{ql}{c@ransac@jointangle}$^\circ$ apart; a typical case and failures of \method: \cref*{fig:supp_qualitative,fig:supp_failures} in \cref*{sec:supp_qualitative}.}
  \label{fig:qualitative}
\end{figure*}

\Cref{tab:panoptic} (top) evaluates every method on identical camera subsets of the 31 Panoptic cameras (camera-count sweep in \cref*{fig:camera_count}, per-subject and per-joint breakdowns in \cref*{tab:supp_subjects,tab:supp_joints}, all in \cref*{sec:supp_panoptic} of the supplement).
\vspace{-2 mm}
\paragraph{Two cameras.}
With two cameras no method can identify a bad view; \method also uses each view's depth, through $\lambda$.
Against DLT it has the lower PA-MPJPE on \pnum{2}{laf}{pa}{beatsransac} of the \pinfo{2}{subsets} pairs and the lower MPJPE on \pnum{2}{laf}{mpjpe}{beatsransac}, and its worst pair has \pnum{2}{laf}{mpjpe}{worst}\,mm MPJPE against \pnum{2}{ransac}{mpjpe}{worst}\,mm.
This gap comes from gross failures: DLT and \method misplace a similar small fraction of joints by more than \gnum{fa}{gross}\,mm (\gnum{fa}{2@ransac@pct}\% and \gnum{fa}{2@laf@pct}\%), but a near-degenerate DLT solve can send a joint up to \gnum{fa}{2@ransac@max}\,m away (\gnum{ql}{a@dlt@jointm}\,m in \cref{fig:qualitative}a), whereas the errors of \method stay within \gnum{fa}{2@laf@max}\,m; refining DLT by minimising the reprojection error does not prevent this (worst pair \pnum{2}{dlt_lm}{mpjpe}{worst}\,mm).
With $\lambda=0$, \method is midpoint triangulation, which fails less badly (worst pair \pnum{2}{midpoint}{mpjpe}{worst}\,mm, joints up to \gnum{gs}{2@midpoint@max}\,m away); the depth term lowers the worst pair to \pnum{2}{laf}{mpjpe}{worst}\,mm.
Triangulation fails worst on the pairs whose viewing rays are within \gnum{pg}{narrowhi}$^\circ$ of collinear: \gnum{pg}{narrowpct}\% of the pairs, but \gnum{pg}{top20narrow} of the 20 worst.
\vspace{-2 mm}
\paragraph{Three and four cameras.}
With a third camera RANSAC can reject a view, and it becomes the strongest baseline in the median; refining its solution by minimising the reprojection error does not help (RANSAC+LM: \pnum{3}{ransac_lm}{mpjpe}{median} vs.\ \pnum{3}{ransac}{mpjpe}{median}\,mm with three cameras, \pnum{4}{ransac_lm}{mpjpe}{median} vs.\ \pnum{4}{ransac}{mpjpe}{median}\,mm with four).
DLT+IRLS cannot reject a view here, because it reverts to all views when fewer than four would remain; its worst case reaches \pnum{3}{dlt_irls}{mpjpe}{worst}\,mm with three cameras and \pnum{4}{dlt_irls}{mpjpe}{worst}\,mm with four, from a few joints placed up to \gnum{fa}{3@dlt_irls@max}\,m away.
\method never discards a view; it down-weights views point by point (\cref{fig:qualitative}b).
Its median error is close to RANSAC's (\pnum{3}{laf}{mpjpe}{median} vs.\ \pnum{3}{ransac}{mpjpe}{median}\,mm MPJPE with three cameras, \pnum{4}{laf}{mpjpe}{median} vs.\ \pnum{4}{ransac}{mpjpe}{median} with four), it has the lower PA-MPJPE on \pnum{3}{laf}{pa}{beatsransac} of \pinfo{3}{subsets} three-camera and \pnum{4}{laf}{pa}{beatsransac} of \pinfo{4}{subsets} four-camera subsets, and its worst-case MPJPE is the lowest of all methods: \pnum{3}{laf}{mpjpe}{worst}\,mm with three cameras and \pnum{4}{laf}{mpjpe}{worst}\,mm with four, against \pnum{3}{ransac}{mpjpe}{worst} and \pnum{4}{ransac}{mpjpe}{worst}\,mm for RANSAC and \pnum{3}{midpoint}{mpjpe}{worst} and \pnum{4}{midpoint}{mpjpe}{worst}\,mm for midpoint triangulation.
RANSAC's threshold is not the reason: leave-one-subject-out picks it from $\{\gnum{rt}{taus}\}$\,px (\gnum{rt}{3@picks}\,px in every fold with three cameras, \gnum{rt}{4@picks}\,px with four), and no threshold brings its worst case below \gnum{rt}{3@minworst}\,mm with three cameras or \gnum{rt}{4@minworst}\,mm with four (held-out: \gnum{rt}{3@heldout@median} / \gnum{rt}{3@heldout@worst} and \gnum{rt}{4@heldout@median} / \gnum{rt}{4@heldout@worst}\,mm).
The failures also differ in kind: RANSAC misplaces fewer joints by more than \gnum{fa}{gross}\,mm than \method (\gnum{fa}{3@ransac@pct}\% vs.\ \gnum{fa}{3@laf@pct}\% with three cameras), but can place them up to \gnum{fa}{3@ransac@max}\,m away, while those of \method stay within \gnum{fa}{3@laf@max}\,m (\cref{fig:qualitative}c with four cameras, \cref*{fig:supp_qualitative}e in \cref*{sec:supp_qualitative} with three; a typical frame, where all methods agree, is \cref*{fig:supp_qualitative}f).
\vspace{-2.5 mm}
\paragraph{A 2D detector pipeline.}
Triangulating the keypoints of a 2D pose detector, YOLOv8x-pose, by DLT, confidence-weighted DLT~\cite{iskakov2019learnable} or RANSAC is the standard multi-view pipeline (\cref*{tab:supp_detector}).
Its keypoints are as accurate in 2D as those of SAM~3D~Body (median \gnum{dt}{px} against \gnum{sd}{median}\,px), but it drops \gnum{dt}{missing}\% of them below its 0.5 confidence, among them \gnum{dt}{missing@ankle}\% of the ankles, which lie outside the image in \gnum{dt}{outside@ankle}\% of the views; we score its triangulations on the joints they produce (coverage \gnum{dt}{2@coverage}--\gnum{dt}{4@coverage}\%), which favours them.
Even so, its best variant has a median MPJPE of \gnum{dt}{2@best@mpjpe@median}, \gnum{dt}{3@best@mpjpe@median} and \gnum{dt}{4@best@mpjpe@median}\,mm with two, three and four cameras, against \pnum{2}{laf}{mpjpe}{median}, \pnum{3}{laf}{mpjpe}{median} and \pnum{4}{laf}{mpjpe}{median}\,mm for \method on every joint, and a far higher worst case (\gnum{dt}{2@best@mpjpe@worst}, \gnum{dt}{3@best@mpjpe@worst} and \gnum{dt}{4@best@mpjpe@worst}\,mm against \pnum{2}{laf}{mpjpe}{worst}, \pnum{3}{laf}{mpjpe}{worst} and \pnum{4}{laf}{mpjpe}{worst}\,mm); confidence weighting changes the median by less than a millimetre.
\vspace{-3 mm}
\paragraph{Selection and weighting are complementary.}
RANSAC decides which views to use at each point; \method decides how far to trust each of them, and in which direction.
The hybrid has the lowest median error in both metrics with three and four cameras, beating RANSAC on the same subset in \pnum{3}{hybrid}{mpjpe}{beatsransac} of \pinfo{3}{subsets} three-camera subsets in MPJPE and \pnum{3}{hybrid}{pa}{beatsransac} in PA-MPJPE (\pnum{4}{hybrid}{mpjpe}{beatsransac} and \pnum{4}{hybrid}{pa}{beatsransac} with four cameras; the gain is not a better refit of the set, \cref*{sec:supp_baselines}).
It inherits RANSAC's failures, since weights cannot repair a wrong consensus set: its worst case is higher than that of \method (Spearman $\rho=\pnum{3}{hybrid}{mpjpe}{rhoransac}$ with RANSAC over three-camera subsets, against \pnum{3}{laf}{mpjpe}{rhoransac} for \method), although up to the 95th percentile it is the more accurate (\pnum{3}{hybrid}{mpjpe}{p95} vs.\ \pnum{3}{laf}{mpjpe}{p95}\,mm with three cameras, \cref*{fig:camera_count} in \cref*{sec:supp_panoptic}).
\vspace{-3 mm}
\paragraph{More cameras.}
With more cameras every robust method improves and the differences shrink (\cref*{fig:camera_count} in \cref*{sec:supp_panoptic}): the hybrid has the lowest median MPJPE at every count from three to 31, and from six cameras on \method alone also has a lower median than every baseline, \eg \pnum{20}{laf}{mpjpe}{median} against \pnum{20}{ransac}{mpjpe}{median}\,mm for RANSAC with 20 cameras.
\vspace{-2 mm}
\paragraph{Parameters.}
Leave-one-subject-out picks the same setting in every fold: \pinfo{2}{laf@picks} with two cameras and \pinfo{3}{laf@picks} with three and four.
With $\lambda=0$ the fused point and the lateral residuals depend only on the viewing rays: \method then uses the same evidence as triangulation, and its advantage comes from how it combines the rays; the per-view depth matters only where the rays alone are ill-conditioned, as with two cameras.
The median is insensitive to the setting (\cref*{sec:supp_baselines}).
\vspace{-3 mm}
\paragraph{A second sequence, nothing tuned.}
On a second Panoptic sequence (another day, \gnum{p2}{subjects} people, 31 cameras), with the settings of the first, the picture is the same: the worst case of \method is below that of every triangulation baseline at every count, its median is slightly above RANSAC's with three and four cameras, and the hybrid has the lowest median there (\cref*{tab:supp_seq2,sec:supp_seq2} in the supplement).

\subsection{Camera placement and count on Fit3D}
\label{sec:exp_fit3d}

Fit3D~\cite{fieraru2021aifit} records fitness exercises with four synchronised cameras in the corners of the capture area, \gnum{fr}{dist@min}--\gnum{fr}{dist@max}\,m from the subject, and a reference from marker-based motion capture.
We use its \gnum{fr}{subjects} training subjects (\gnum{fr}{frames} frames) and score the 12 limb joints that both skeletons define (\cref*{sec:supp_fit3d}).
Nothing is tuned on Fit3D: \method and the hybrid reuse the setting selected on Panoptic for each camera count, and RANSAC keeps its 20\,px.
\Cref{tab:panoptic} (bottom) shows the results.
\vspace{-2 mm}
\paragraph{Two cameras.}
Triangulation fails on the two pairs of cameras that face each other across the room, whose rays are within \gnum{fr}{bad@cross}$^\circ$ of collinear: DLT reaches \gnum{fr}{bad@dlt@min}--\gnum{fr}{bad@dlt@max}\,mm MPJPE there, \method \gnum{fr}{bad@laf@min}--\gnum{fr}{bad@laf@max}\,mm.
\method thus lowers the worst pair from \xnum{fr}{2}{dlt_irls}{mpjpe}{worst} to \xnum{fr}{2}{laf}{mpjpe}{worst}\,mm, below midpoint triangulation as well (\xnum{fr}{2}{midpoint}{mpjpe}{worst}\,mm), and on the other four pairs it is at most \gnum{fr}{good@lafgap@max}\,mm worse than DLT.
\vspace{-2.5 mm}
\paragraph{Three and four cameras.}
These subsets see the subject from well-separated directions, and all methods agree within \xinfo{fr}{34}{spread}\,mm in median and worst case; RANSAC rejects only \xinfo{fr}{3}{rejected}\% of the camera-points.
Errors are higher than on Panoptic mainly because MPJPE here also contains the difference between the COCO keypoints of SAM~3D~Body and the anatomical joint centres of Fit3D (\cref*{sec:supp_fit3d}).

\subsection{Other estimators and ablation}
\label{sec:exp_estimator}\label{sec:exp_ablation}

\paragraph{Other per-view estimators.}
We repeat the camera study with HMR~2.0~\cite{goel2023humans} and CameraHMR~\cite{patel2025camerahmr} in place of SAM~3D~Body, with the same frames, boxes, subsets and parameter grid; Panoptic is not in the training data of either (\cref*{tab:hmr2,tab:supp_hmr2,tab:supp_camerahmr_fov,tab:supp_camerahmr,sec:supp_hmr2} in the supplement).
Both are less accurate (single-view MPJPE \gnum{es}{hm@single@mpjpe} and \gnum{es}{cf@single@mpjpe} against \gnum{es}{sam@single@mpjpe}\,mm after \cref{eq:shift}).
The main conclusion of \cref{sec:exp_panoptic} holds for both: \method has a lower worst case than every triangulation method with two, three and four cameras, and with HMR~2.0 also a lower median than every baseline; with CameraHMR, whose keypoints are nearly as precise as SAM~3D~Body's, the pattern of \cref{sec:exp_panoptic} returns.
\vspace{-3 mm}
\paragraph{Ablation.}
\Cref*{tab:ablation} in the supplement evaluates the contribution of each component (variants defined in \cref*{sec:supp_baselines}).
Anisotropic precision is essential: with uniform weights, depth-corrected estimates achieve \gnum{ab}{3@uniform@mpjpe}\,mm MPJPE using three cameras, while agreement-based weights with isotropic precision ($\lambda=1$) yield \gnum{ab}{3@isotropic@mpjpe}\,mm, worse than the unweighted mean (\gnum{ab}{3@mean@mpjpe}\,mm with depth correction; \gnum{ab}{3@mean_raw@mpjpe}\,mm without).
Computing agreement perpendicular to the viewing ray further improves accuracy over using the full residual: \gnum{ab}{3@laf@mpjpe} versus \gnum{ab}{3@full_residual@mpjpe}\,mm with three cameras, and \gnum{ab}{5@laf@mpjpe} versus \gnum{ab}{5@full_residual@mpjpe}\,mm with five.
This supports the central design principle of \method: along-ray residuals are dominated by monocular depth errors and are therefore unreliable indicators of view consistency.
\vspace{-2 mm}
\paragraph{Runtime.}
Fusing the \gnum{tm}{points} points of a frame takes \method \gnum{tm}{4@laf}\,ms with four cameras and \gnum{tm}{31@laf}\,ms with 31 (NumPy, \gnum{tm}{cpu}), \gnum{tm}{4@dlt_irls@ratio} and \gnum{tm}{31@dlt_irls@ratio} times faster than DLT+IRLS; RANSAC and the hybrid take \gnum{tm}{4@ransac} and \gnum{tm}{31@ransac}\,ms (\cref*{sec:exp_runtime}).
Per-view inference takes \gnum{ti}{sperframe}\,s per camera and frame on a \gnum{ti}{gpu}, the same for every method.

%% file: tables/combined_main.tex
\begin{tabular}{@{}lcccc@{\hskip 10pt}cccc@{\hskip 10pt}cccc@{}}
\toprule
 & \multicolumn{4}{c}{2 cameras} & \multicolumn{4}{c}{3 cameras} & \multicolumn{4}{c}{4 cameras}\\
\cmidrule(lr){2-5}\cmidrule(lr){6-9}\cmidrule(lr){10-13}
 & \multicolumn{2}{c}{MPJPE} & \multicolumn{2}{c}{PA-MPJPE} & \multicolumn{2}{c}{MPJPE} & \multicolumn{2}{c}{PA-MPJPE} & \multicolumn{2}{c}{MPJPE} & \multicolumn{2}{c}{PA-MPJPE}\\
Method & med. & worst & med. & worst & med. & worst & med. & worst & med. & worst & med. & worst\\
\midrule
\multicolumn{1}{@{}>{\columncolor{gray!15}[0pt][\tabcolsep]}l}{\textit{CMU Panoptic}} & \multicolumn{4}{>{\columncolor{gray!15}}c}{\textit{all 465}} & \multicolumn{4}{>{\columncolor{gray!15}}c}{\textit{1{,}000 of 4{,}495}} & \multicolumn{4}{>{\columncolor{gray!15}}c}{\textit{1{,}000 of 31{,}465}}\\
DLT+IRLS & 30.0 & 284.3 & 23.0 & 104.7 & 26.9 & 178.1 & 20.7 & 33.7 & 25.7 & 467.0 & 19.8 & 28.9\\
IRLS-2 & 30.0 & 284.3 & 23.0 & 104.7 & 26.8 & 178.1 & 20.7 & 33.7 & 25.3 & 466.3 & 19.9 & 28.8\\
DLT+LM & 29.9 & 273.4 & 23.0 & 100.5 & 27.1 & 73.3 & 20.7 & 32.7 & 26.0 & 246.8 & 20.0 & 28.0\\
Midpoint & 29.6 & 88.2 & 22.9 & 91.5 & 26.6 & 39.8 & 20.5 & 32.5 & 25.3 & 35.1 & 19.6 & 28.1\\
RANSAC & 30.0 & 284.3 & 23.0 & 104.7 & 24.9 & 61.8 & 20.3 & 47.7 & 22.9 & 45.5 & 18.6 & 37.9\\
RANSAC+LM & 29.9 & 273.4 & 23.0 & 100.5 & 25.0 & 62.5 & 20.3 & 47.6 & 23.0 & 45.1 & 18.7 & 37.8\\
\midrule
\method, $\lambda{=}0.02$, $\alpha{=}0.5$ & 29.7 & \textbf{52.5} & \textbf{22.3} & \textbf{37.4} & 26.2 & 38.9 & 20.0 & \textbf{28.7} & 23.8 & 33.5 & 18.4 & \textbf{25.1}\\
\method, held-out & \textbf{29.4} & 56.0 & 22.5 & 44.2 & 25.2 & \textbf{37.8} & 20.0 & 30.3 & 22.8 & \textbf{32.3} & 18.3 & 25.4\\
RANSAC set + \method, held-out & \textbf{29.4} & 56.0 & 22.5 & 44.2 & \textbf{24.6} & 43.9 & \textbf{19.9} & 43.4 & \textbf{22.4} & 37.4 & \textbf{18.1} & 35.9\\
\midrule
\multicolumn{1}{@{}>{\columncolor{gray!15}[0pt][\tabcolsep]}l}{\textit{Fit3D}} & \multicolumn{4}{>{\columncolor{gray!15}}c}{\textit{all 6}} & \multicolumn{4}{>{\columncolor{gray!15}}c}{\textit{all 4}} & \multicolumn{4}{>{\columncolor{gray!15}}c}{\textit{single subset}}\\
DLT+IRLS & 57.3 & 95.8 & 47.7 & 69.3 & 55.6 & 56.3 & 46.5 & 47.1 & 55.0 & 55.0 & 46.0 & 46.0\\
IRLS-2 & 57.3 & 95.8 & 47.7 & 69.3 & 55.6 & 56.3 & 46.5 & 47.1 & 55.0 & 55.0 & 46.0 & 46.0\\
DLT+LM & \textbf{57.2} & 94.9 & 47.6 & 68.3 & 55.4 & \textbf{56.1} & 46.3 & 46.9 & 54.7 & 54.7 & 45.7 & 45.7\\
Midpoint & \textbf{57.2} & 77.7 & 47.7 & 67.7 & 55.6 & 56.3 & 46.5 & 47.1 & 55.0 & 55.0 & 46.0 & 46.0\\
RANSAC & 57.3 & 95.8 & 47.7 & 69.3 & 55.5 & 56.2 & 46.5 & 47.2 & 54.5 & 54.5 & 45.6 & 45.6\\
RANSAC+LM & \textbf{57.2} & 94.9 & 47.6 & 68.3 & 55.4 & \textbf{56.1} & 46.4 & 47.1 & \textbf{54.3} & \textbf{54.3} & \textbf{45.4} & \textbf{45.4}\\
\midrule
\method, held-out & 58.5 & \textbf{65.4} & \textbf{47.4} & \textbf{52.2} & \textbf{55.3} & \textbf{56.1} & \textbf{46.2} & \textbf{46.6} & 54.5 & 54.5 & \textbf{45.4} & \textbf{45.4}\\
RANSAC set + \method, held-out & 58.5 & \textbf{65.4} & \textbf{47.4} & \textbf{52.2} & 55.5 & 56.3 & 46.4 & 46.9 & 54.5 & 54.5 & \textbf{45.4} & \textbf{45.4}\\
\bottomrule
\end{tabular}

%% file: sec/5_conclusion.tex
\vspace{-4mm}
\section{Limitations and Conclusion}
\label{sec:conclusion}

Monocular estimators recover plausible 3D bodies but often misestimate depth. \method exploits the known viewing rays to fuse these estimates without training, using three $3\times3$ solves per point: trust the ray, not the depth. Across two to four cameras, it reduces worst-case error relative to triangulation baselines on Panoptic and avoids two-camera triangulation failures on Fit3D, while remaining competitive with more cameras. \method requires calibrated cameras and consistent mesh topology. It fuses points independently, without temporal or body-shaped priors. Evaluation is limited to two studios, nine Panoptic subjects, and eight Fit3D subjects, with no multi-person scenes. Its median error is comparable to that of RANSAC's, while the hybrid improves the median but inherits RANSAC's worst cases. 